\documentclass[letterpaper]{article} 
\usepackage{aaai2026}  
\usepackage{times}  
\usepackage{helvet}  
\usepackage{courier}  
\usepackage[hyphens]{url}  
\usepackage{graphicx} 
\usepackage{natbib}  
\usepackage{caption} 
\usepackage{algorithm}
\usepackage{algorithmic}
\usepackage{booktabs}       
\usepackage{amsmath}
\usepackage{amsfonts}       
\usepackage{amssymb}  
\usepackage{multirow}
\usepackage{newfloat}
\usepackage{listings}
\DeclareCaptionStyle{ruled}{labelfont=normalfont,labelsep=colon,strut=off} 
\floatstyle{ruled}
\newfloat{listing}{tb}{lst}{}
\floatname{listing}{Listing}
\usepackage{listings} 
\usepackage{xcolor}   
\usepackage{enumitem} 
\usepackage{listings}

\usepackage{listings}

\definecolor{myred}{RGB}{220, 50, 47}

\lstdefinestyle{verilog}{
    language=Verilog,
    basicstyle=\ttfamily\footnotesize,
    keywordstyle=\color{blue},
    commentstyle=\color{gray},
    stringstyle=\color{orange},
    showstringspaces=false,
    breaklines=true,
    frame=none,
    moredelim=**[is][\color{myred}]{@}{@}, 
}

\title{QiMeng-CRUX: Narrowing the Gap Between Natural Language and Verilog via Core Refined Understanding eXpression for Circuit Design }

\author {
    Lei Huang\textsuperscript{\rm 1,\rm 2},
    Rui Zhang\textsuperscript{\rm 1},
    Jiaming Guo\textsuperscript{\rm 1},
    Yang Zhang\textsuperscript{\rm 1,\rm 2},
    Di Huang\textsuperscript{\rm 1},
    Shuyao Cheng\textsuperscript{\rm 1}, \\
    Pengwei Jin\textsuperscript{\rm 1,\rm 2},
    Chongxiao Li\textsuperscript{\rm 1,\rm 2},
    Zidong Du\textsuperscript{\rm 1},
    Xing Hu\textsuperscript{\rm 1},
    Yunji Chen\textsuperscript{\rm 1,\rm 2},
    Qi Guo\textsuperscript{\rm 1}\thanks{Corresponding author. Contact:guoqi@ict.ac.cn}
}

\affiliations{
    \textsuperscript{\rm 1}State Key Lab of Processors, Institute of Computing Technology, CAS
    
    \textsuperscript{\rm 2}University of Chinese Academy of Sciences

}

\usepackage{bibentry}

\begin{document}

\maketitle

\begin{abstract}

Large language models (LLMs) have shown promising capabilities in hardware description language (HDL) generation. However, existing approaches often rely on free-form natural language descriptions that are often ambiguous, redundant, and unstructured, which poses significant challenges for downstream Verilog code generation. 
We treat hardware code generation as a complex transformation from an open-ended natural language space to a domain-specific, highly constrained target space. 
To bridge this gap, we introduce \textbf{C}ore \textbf{R}efined \textbf{U}nderstanding e\textbf{X}pression (CRUX), a structured intermediate space that captures the essential semantics of user intent while organizing the expression for precise Verilog code generation.
We further design a two-stage training framework, comprising \textit{Joint Expression Modeling} and \textit{Dual-Space Optimization}, to enhance the quality of both CRUX and Verilog code.
Experiments across multiple Verilog generation benchmarks demonstrate that our model, QiMeng-CRUX, achieves state-of-the-art performance among general models, particularly under challenging design tasks. Furthermore, the CRUX space proves transferable and beneficial when used as input prompts for other code models, highlighting its effectiveness in narrowing the gap between free-form natural language descriptions and precise Verilog generation.

\end{abstract}


\begin{figure}[!t]
    \captionsetup{skip=5pt, belowskip=-10pt}
    \centering
    \hspace{-0.3cm} 
    \includegraphics[width=0.48\textwidth]{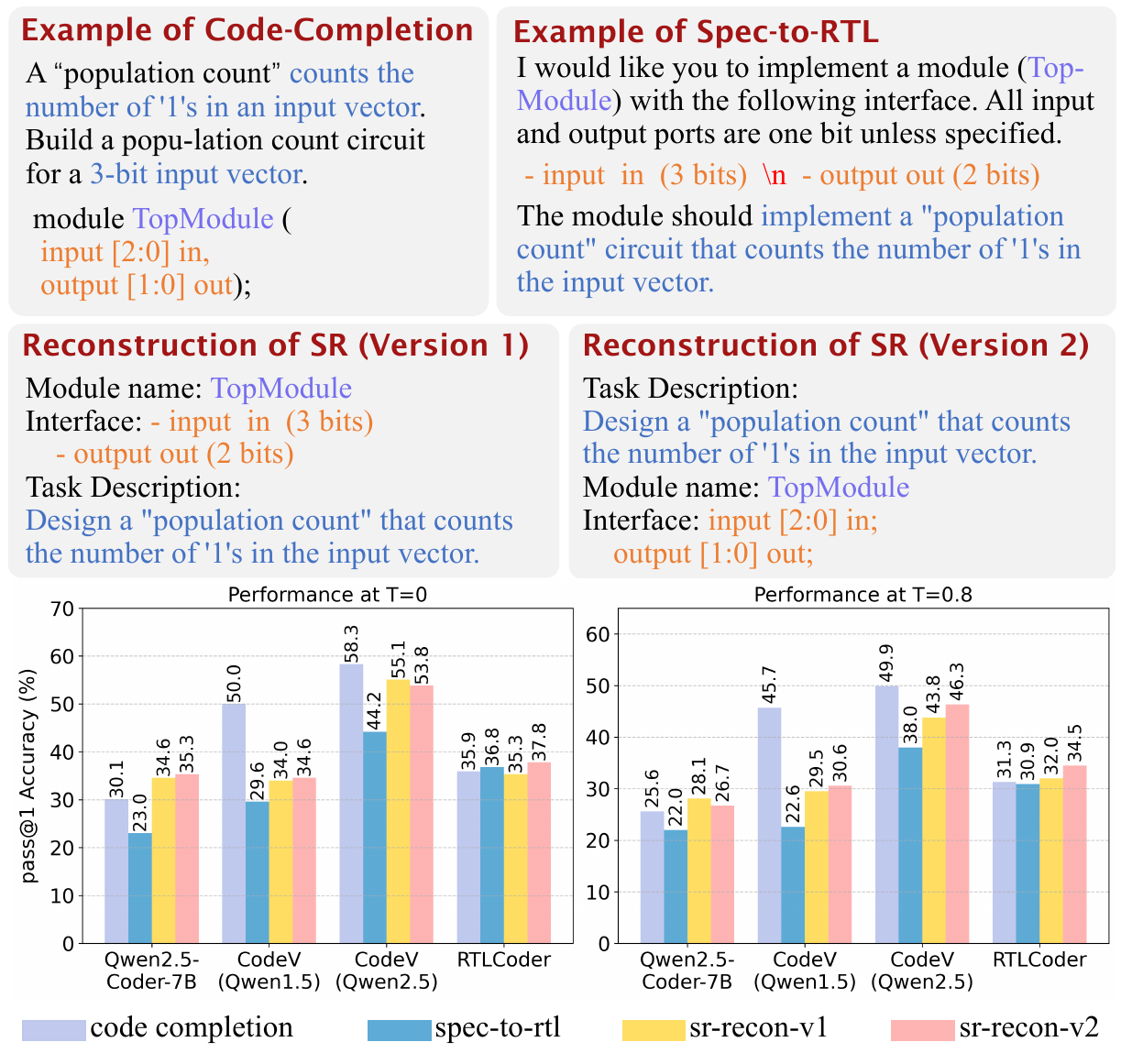}
    \caption{Descriptions of the Code-Completion (CC) and Spec-to-RTL (SR) tasks in the VerilogEval-v2 benchmark convey equivalent design intent, but differ in structure. The top portion shows an example, while the bottom presents the performance of several general-purpose and domain-specific code LLMs on the whole benchmark. Results show that expression structure significantly impacts model performance, with SR-Recon consistently outperforming SR, and CC benefiting CodeVs under more constrained input formats.}

    \label{fig:analysis}
\end{figure}

\begin{figure*}[!t]
    \captionsetup{skip=2pt, belowskip=-10pt}
  \centering
  \includegraphics[width=\textwidth]{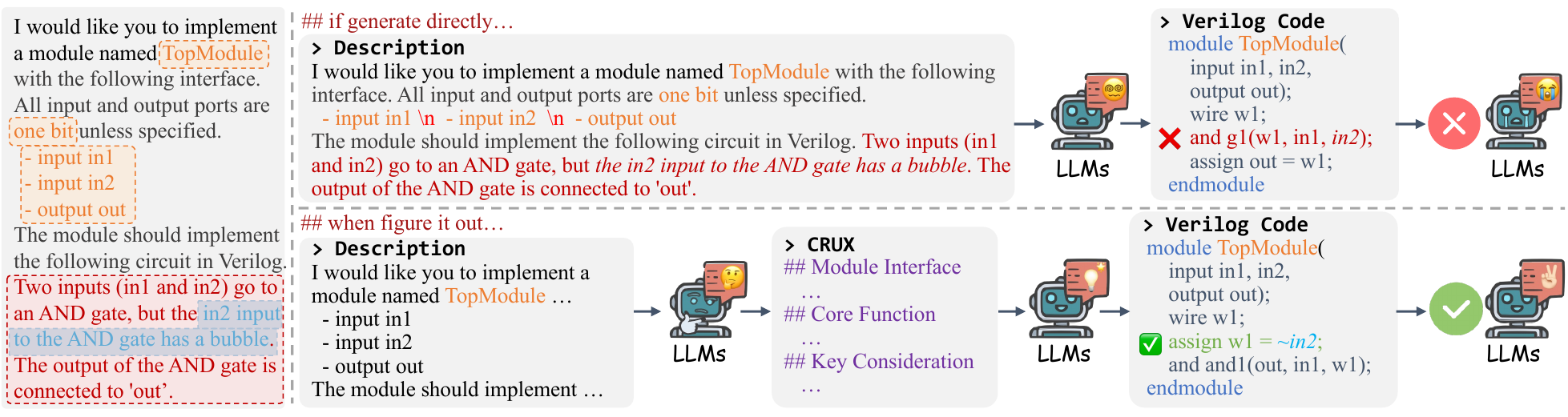} 
  \caption{
  An example from the VerilogEval-v2 Spec-to-RTL \cite{verilogeval-v2} benchmark shows that the original description lacks explicit emphasis on critical design details. Feeding this directly to LLMs often leads to incorrect implementations due to misinterpretation of key details. While the model possesses the capability to generate correct code, the ambiguity and underspecification in the input hinder accurate realization. In contrast, providing the model with CRUX enables more reliable alignment with the desired design specification.
  }
  \label{fig:demonstration}
  \vspace{-3pt}
\end{figure*}

\section{Introduction}

Large language models (LLMs) have shown remarkable progress in code generation for general-purpose programming languages, spurring increasing interest in automating hardware description language (HDL) code generation, such as Verilog \cite{dawn}. 
Recent efforts have substantially improved LLM performance on hardware design tasks through advancements in many aspects, such as data construction \cite{rtlcoder,codev,mg-verilog}, feedback mechanisms \cite{insightsfromverification, origen, codev-r1}, and training paradigms \cite{aotovcoder, betterv}. 
However, most existing LLM-based methods for Verilog generation focus primarily on the code generation phase and rely solely on free-form natural language descriptions, with limited attention to how the expression of descriptions influences downstream performance.

In practice, free-form natural language descriptions of hardware design from users often suffer from ambiguity in intent, redundancy in expression, and irregularity in structure, particularly in critical modules involving sequential logic, finite state machines (FSMs), and complex control behaviors \cite{craftrtl}. Thus, crucial information is often scattered, underspecified, or only implicitly conveyed through the text, posing significant challenges for the model to organize and understand the necessary semantics. In contrast, Verilog is a formal specification language for modeling hardware circuits, characterized by strict structural constraints, strong semantic rigor, and high demands for domain expertise \cite{betterv}. 
Such gap reveals a fundamental challenge in Verilog code generation task: transforming from a highly expressive and inherently unstructured description space into a structured code space governed by syntactic rigidity, modular hierarchy, and functional alignment. The high difficulty of this transformation poses significant obstacles to existing LLM-based approaches, often resulting in misaligned design intent, semantic drift, and distorted implementations in engineering.

Motivated by this gap, we investigate the effect of input formulation and find semantically structured and refined expressions leading to better generation performance. 
As illustrated in Figure \ref{fig:analysis}, even when the descriptions of the two tasks contain the same underlying information, the model's performance can vary substantially depending on how the information is expressed. Notably, when the original input of Spec-to-RTL \cite{verilogeval-v2} is restructured to expose its key components more explicitly and precisely, the models demonstrate significant improvements in generation accuracy, suggesting the importance of input clarity and organization in guiding LLMs toward robust code generation. 

Based on the analysis above, we propose \textbf{C}ore \textbf{R}efined \textbf{U}nderstanding e\textbf{X}pression (CRUX), which serves as a structured intermediate space that narrows the gap between free-form descriptions and formal Verilog implementations. The CRUX space captures the essential semantics of the user intent while structuring the descriptions for Verilog code generation, offering a stable and precise foundation for producing synthesizable and correct implementations. 
CRUX comprises three key components: \textit{Module Interface}, \textit{Core Functions}, and \textit{Key Considerations}, each addressing a distinct aspect of Verilog generation. Specifically, 
(i) The \textit{Module Interfaces} establish the structural foundation for interface-compliant Verilog modules by explicitly defining the input and output ports along with their signal properties.
(ii) The \textit{Core Functions} capture the essential circuit behaviour logic, clearly specifying functional goals and guiding the overall control and data flow. 
(iii) The \textit{Key Considerations} highlight subtle but critical implementation details and constraints to ensure the generated code is both synthesizable and accurate.
Building on these components, CRUX provides a coherent engineering context and functional blueprint for hardware design, enabling precise code generation with respect to structural completeness, behavioral fidelity, and synthesis feasibility, as shown in Figure~\ref{fig:demonstration}.

\begin{figure*}[!t] 
    \captionsetup{skip=2pt, belowskip=-10pt}
  \centering
  \includegraphics[width=1.02\textwidth]{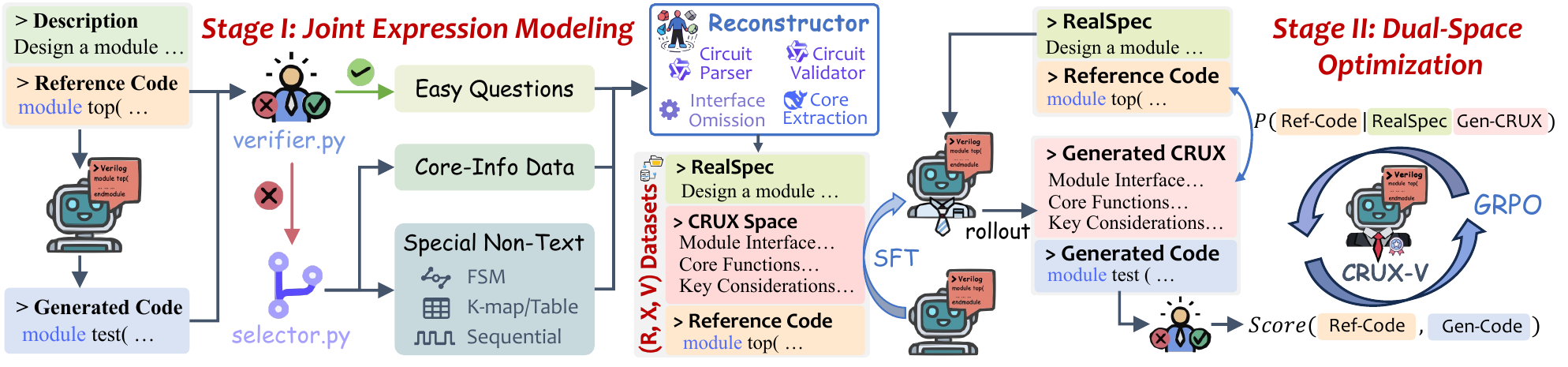} 
  \caption{
  Overview of the two-stage training process. Stage I involves dataset categorization and reconstruction, which is used for supervised fine-tuning (SFT). Stage II applies CRUX-Enhanced GRPO for RL-based post-training.
  }
  \label{fig:overview}
  \vspace{-3pt}
\end{figure*}

To fully leverage the guiding potential of CRUX, we develop a comprehensive framework composed of two stages: \textit{Joint Expression Modeling} and \textit{Dual-Space Optimization}, as shown in Figure \ref{fig:overview}.
In the first stage, we construct the CRUX through a combination of LLM-based generation and category-specific strategies, over a diverse set of descriptions reflecting real-world human usage. We then perform supervised fine-tuning (SFT) using both CRUX and Verilog code as supervision signals, enabling the model to establish an initial mapping from natural language descriptions to structured CRUX expressions, as well as the corresponding code generation capabilities.
In the second stage, we introduce CRUX-enhanced GRPO, an RL-based procedure that jointly optimizes two interconnected spaces: the CRUX space and the Verilog code space. By going beyond the limitations of static supervision, this dual-space optimization enables the model to adaptively converge toward high-quality solution regions aligned with target design intents.

Comprehensive experiments on multiple Verilog generation benchmarks demonstrate that our model QiMeng-CRUX, achieves state-of-the-art performance across both general-purpose and Verilog-specific models, especially under challenging and realistic design scenarios like Spec-to-RTL (from 49.3\% to 64.7\% with T=0, from 46.8\% to 64.4\% with T=0.8) and RTLLM-v2 (from 50.9\% to 63.8\%). In addition, the CRUX space proves to be a robust and semantically meaningful intermediate guidance that enhances performance even when used as a prompt for other code models without any further training. These findings highlight the potential of CRUX as a powerful and effective approach for supporting precise Verilog code generation.
Code is available at https://github.com/Taskii-Lei/QiMeng-CRUX-V.

\section{Related Work}
LLM-based Verilog code generation has gained increasing attention as researchers explore the potential of LLMs for automated hardware design. Rather than relying solely on general-purpose techniques, recent efforts seek to adapt and extend generation paradigms to accommodate the characteristics unique to Verilog. These explorations have spanned multiple directions, including dataset construction, framework design, feedback-driven optimization and so on.

\paragraph{Dataset Construction and Enhancement} 
High-quality datasets play a central role in Verilog code generation. Early works such as Dave \cite{pearce2020dave} explored fine-tuning of GPT-2 to generate Verilog snippets from natural language, marking one of the first attempts to align LLMs with hardware tasks. Recently, several works mined codes from public sources and refined task descriptions \cite{rtlcoder, verigen, codev}. CraftRTL \cite{craftrtl} extended this by introducing non-text tasks like FSM to increase diversity. Other efforts including hdl2v \cite{hdl2v} and MG-Verilog \cite{mg-verilog}, focused on improving data quality through code translation and fine-grained annotation. These datasets collectively enhance both the scale and quality of Verilog training data.

\paragraph{Optimization via Feedback}
Given that Verilog code is ultimately executed and verified by downstream EDA tools, recent studies have explored two main strategies for incorporating feedback to improve generation quality.
One line of work applies feedback at the inference stage, where either verification feedback \cite{insightsfromverification, automaticallyimproving, romanwasnotbuilt}, or model feedback \cite{origen, vrank} is used to identify and revise incorrect outputs.  
The other line leverages RL-based post-training \cite{schulman2017ppo,bai2022RLHF,shao2024grpo,yu2025dapo} to optimize generation with reward functions \cite{coderl, dou2024stepcoder, veriseek, codev-r1}, which may incorporate verification feedback, structural constraints, and design priors. This enables models to move beyond static supervision and align more closely with hardware-centric design objectives.

\paragraph{CoT Reasoning}
Motivated by the success of DeepSeek-R1 \cite{guo2025deepseek}, which demonstrates the benefits of chain-of-thought (CoT) reasoning with RL, recent studies have tried to incorporate such diagram into Verilog code generation with verification feedback. Most CoT supervision is obtained through distillation from stronger models \cite{verithoughts, reasoningv,codev-r1}, while \citeauthor{verireason} adopts optimization-based pipelines to construct thinking trajectories. While effective, CoT-based methods often require higher training costs and longer inference time compared to general models \cite{haven}.

\section{Methodology}
Our goal is to generate structurally and functionally correct Verilog code from natural language descriptions by narrowing the gap between natural language descriptions and the Verilog code domain. 
To this end, we leverage CRUX as a domain-oriented intermediate space that explicitly captures core design intent and reformulates free-form descriptions into a domain-aligned structural form better supporting downstream code generation. We develop a two-stage training framework to support this process: \textit{Joint Expression Modeling}, which jointly learns to generate both CRUX and Verilog from natural language descriptions, and \textit{Dual-Space Optimization}, which further refines the CRUX and code quality through reinforcement learning in both spaces.

\subsection{Joint Expression Modeling}
To equip the model with the ability to construct CRUX and generate Verilog code, we begin with supervised fine-tuning (SFT). However, most existing datasets are synthesized and exhibit limited diversity in descriptions. So we augment the dataset of CodeV \cite{codev} by simulating real-world user phrasing via introducing variations, ambiguities, and structural incompleteness. Then we derive corresponding CRUX space to explicitly extract and organize the core design intent, resulting in a reconstructed dataset of triplets (R, X, V), where R denotes the realistic task specifications (RealSpec), X is the CRUX, and V is the target Verilog code implementations. Finally, we perform SFT using both CRUX and Verilog code as learning targets. The model is trained to generate the CRUX first from the RealSpec, and then the Verilog code.

\paragraph{Corpus Categorization} 
Given the diverse nature of Verilog tasks, we aim to identify which problems intrinsically require structural guidance beyond plain-text prompts. In particular, we are interested in a subset of circuit-specific tasks that cannot be adequately described or resolved through natural language alone, such as Karnaugh map interpretation, FSM modeling, or waveform analysis \cite{craftrtl}. To systematically explore this heterogeneity, we categorize the dataset into three types: \textit{Easy Questions}, \textit{Special Non-text} and \textit{Normal Data}.
We begin by prompting LLMs with task descriptions to generate corresponding Verilog code. Functional correctness is then automatically verified via testbenches comparing generated outputs with references under identical inputs. Tasks passing verification are labeled Easy Questions. Among failures, those containing keywords such as "k-map," "FSM," or "sequential/waveform" are classified as Special Non-Text, with the rest assigned to Normal Data.

\begin{figure}[!t]
    \captionsetup{skip=5pt, belowskip=-10pt}
    \centering
    \includegraphics[width=0.48\textwidth]{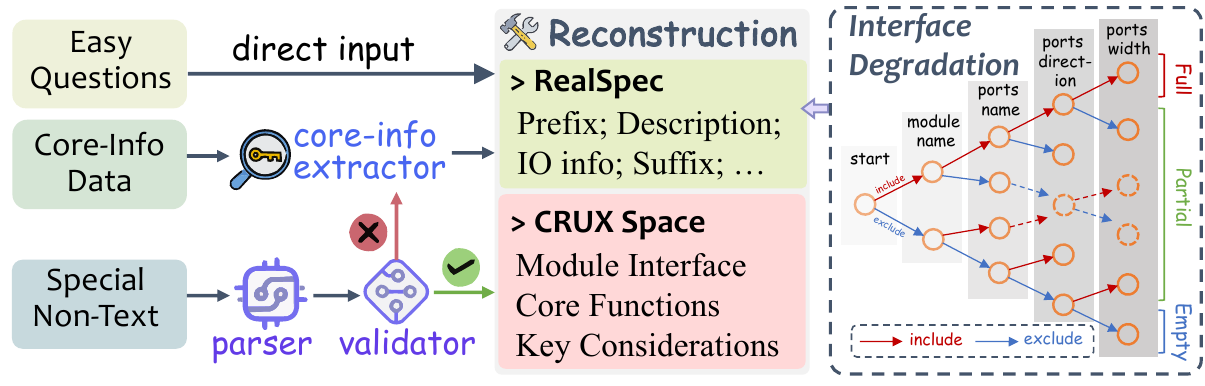}
    \caption{We apply different processing pipelines to the three categories. RealSpec uses prefix/suffix augmentation and interface degradation for variation, while CRUX is constructed mainly via LLMs.}
    \label{fig:reconstruction}
    \vspace{-5pt}
\end{figure}

\paragraph{RealSpec Construction}
RealSpec is designed to simulate natural language descriptions that better reflect realistic distributions. 
While CodeV relies on synthetic prompts generated by ChatGPT, which inherently contain some noise, we introduce further variability to better match real-world conditions.
Specifically, we parse each module’s header to extract detailed interface specifications (e.g., signal names, directions, and bit-widths), then selectively drop or retain parts of this information with a fixed probability, as shown in the right of Figure \ref{fig:reconstruction}. 
The degraded interfaces are inserted into either the middle or the end of the original descriptions at a preset ratio.
Particularly for Special Non-Text, we replace purely textual descriptions (as used in CodeV) with diagrams generated during the CRUX Deviation step, and only append the degraded interface at the end. Finally, we randomly add some prefixes and suffixes to ensure the fluency of the descriptions. 
Unlike CodeV’s fully specified prompts, RealSpec deliberately introduces interface degradation and controlled noise, pushing the model to infer missing components and thus improving robustness in realistic settings.

\paragraph{CRUX Derivation} 
We define the CRUX space as comprising three components: \textit{Module Interface}, \textit{Core Functions}, and \textit{Key Considerations}. The \textit{Module Interface} comes from the reference code header, while the other two are derived differently depending on the corpus type, as illustrated in the left of Figure \ref{fig:reconstruction}.
(i) \textit{Easy Questions}: The descriptions are considered sufficiently clear and complete, so we directly adopt them as \textit{Core Functions}, leaving \textit{Key Considerations} empty.
(ii) \textit{Normal Data}: As the primary source for constructing CRUX, these tasks are processed by prompting DeepSeek-R1 \cite{guo2025deepseek} with the descriptions and reference codes to extract and refine the \textit{Core Functions} and \textit{Key Considerations}.
(iii) \textit{Special Non-Text}: We use Qwen2.5-Coder-32B-Instruct \cite{hui2024qwen2} as a circuit parser to generate diagrams and concise analyses from both the description and code. This includes “State → Condition → Next State” for FSMs, logical expressions for K-maps or truth tables, and core behavior for sequential circuits. The generated diagram and code are then validated by Qwen2.5-Coder-32B-Instruct. Only valid samples are retained to construct the RealSpec and CRUX, while parsing failures or incorrect cases are reclassified into \textit{Normal Data} to process.

\paragraph{Ready and SFT}
After the steps above, we reconstruct the original dataset of CodeV \cite{codev} into (R, X, V) tuples and perform SFT to establish the LLM’s initial capacity for generating CRUX and correct Verilog code based on RealSpec.

\subsection{Dual-Space Optimization}

While SFT enables initial learning from static data, it often yields an under-refined CRUX space that lacks sufficient guidance for precise Verilog generation. So we introduce a Dual-Space Optimization framework that builds upon a CRUX-enhanced GRPO, where the model takes a RealSpec as input and rollouts both the CRUX and the corresponding Verilog implementation. Crucially, we design a composite reward that reflects two interconnected goals: (i) ensuring the generated code behaves correctly, and (ii) promoting CRUX space that better supports code generation.

The first goal is achieved by leveraging Code-Reward, which quantifies the functional correctness of the generated implementation. Specifically, we use the automated verifier to execute both the predicted and reference codes under identical input conditions and compute output matching scores.
These scores reflect how well the generated code adheres to the intended behavior and serve as the basis for assigning reward signals during training.

For the second, rather than aligning the generated CRUX with a reference in a supervised manner, we use CRUX-Reward to measure how much it helps the model confidently arrive at a correct implementation. 
Intuitively, a well-formed CRUX should clearly identify the core design intent and filter out noise, thereby narrowing the gap with a more concentrated distribution centered on valid implementations. 

In practice, we approximate this effect by computing the conditional sequence log-likelihood of the reference code given the RealSpec and the generated CRUX. 
To bound the reward within an interpretable range, we exponentiate the mean log-likelihood, yielding it as a direct measure of the model’s confidence: higher values indicate more concentrated distributions centered on valid implementations. 
Formally, given a training sample $(r, x, v_{ref}) \in (R, X, V)$, the model first rollouts the CRUX $x'$ and then code $c'$ conditioned on the Realspec $r$. The CRUX-Reward is defined as:
\begin{align}
\label{eq:crux_reward}
\text{CRUX-Reward} 
&= \exp\left( \frac{1}{L} \sum_{i=0}^{L-1} \log \pi_{\theta}(y_{i+1} \mid r, x', y_{0:i}) \right) \\
&\text{where } y_i \in v_{\text{ref}}, \quad L = |v_{\text{ref}}| \notag
\end{align}
This contrasts with conventional pipelines that optimize solely for final code correctness, often treating intermediate steps as auxiliary byproducts. Our formulation instead treats CRUX as one of the central modeling targets in the optimization loop, reinforcing its role as a structured intermediate space that bridges the intent and implementation.

\begin{table*}[!t]
\centering
{\fontsize{9}{9.5}\selectfont  
\begin{tabular}{@{}ccccccccccc@{}}
\toprule
\multirow{2}{*}{Type} & \multirow{2}{*}{Model} & \multicolumn{3}{c}{VE-v1-Machine} & \multicolumn{3}{c}{VE-v1-Human} & \multicolumn{3}{c}{RTLLM-V1} \\ \cmidrule(l){3-11} 
 &  & p@1 & p@5 & p@10 & p@1 & p@5 & p@10 & p@1 & p@5 & p@10 \\ \midrule
\multirow{2}{*}{\begin{tabular}[c]{@{}c@{}}Foundation\\ General Models\end{tabular}} & GPT-4o & 67.7 & 75.5 & 77.2 & 60.1 & 71.4 & 74.5 & 41.7 & 65.9 & - \\
 & Deepseek-V3-671B & 77.6 & 86.2 & 87.4 & 70.7 & 77.4 & 78.8 & 60.9 & 74.2 & - \\ \midrule
\multirow{2}{*}{\begin{tabular}[c]{@{}c@{}}Foundation\\ Reasoning Models\end{tabular}} & Deepseek-R1-671B & \textbf{81.0} & \textbf{87.4} & \textbf{89.5} & \textbf{81.5} & \textbf{87.6} & \textbf{88.5} & 64.8 & 82.9 & - \\
 & QWQ-32B & 71.1 & 84.0 & 87.0 & 63.6 & 78.0 & 81.3 & 50.9 & 70.6 &  \\ \midrule
\multirow{3}{*}{\begin{tabular}[c]{@{}c@{}}Verilog-Specific\\ Reasoning Models\end{tabular}} & HaVen-7B & 77.3 & 81.2 & - & 61.1 & 64.8 & - & 62.2 & - & - \\
 & CodeV-R1-Distill-7B & 76.2 & 85.6 & 87.0 & 65.7 & 76.8 & 79.7 & 57.4 & 75.8 & - \\
 & CodeV-R1-7B & 76.5 & 84.1 & 85.7 & 69.9 & 79.3 & 81.9 & \textbf{72.9} & \textbf{86.1} & - \\ \midrule
\multirow{3}{*}{\begin{tabular}[c]{@{}c@{}}General \\ Code Models\end{tabular}} & Qwen2.5-Coder-7B & 50.1 & 66.5 & 70.9 & 22.9 & 36 & 39.5 & 32.2 & 48.2 & 56.0 \\
 & Qwen2.5-Coder-32B & 66.6 & 76.6 & 79.7 & 47.6 & 58.1 & 61.8 & 47.9 & 67.7 & - \\
 & Deepseek-Coder-6.7B & 52.2 & 55.4 & 56.8 & 30.2 & 33.9 & 34.9 & 30.2 & 44.0 & 52.5 \\ \midrule
\multirow{7}{*}{\begin{tabular}[c]{@{}c@{}}Verilog-Specific \\ General Models\end{tabular}} & RTLCoder-6.7B & 61.2 & 76.5 & 81.8 & 41.6 & 50.1 & 53.4 & 35.8 & 40.3 & 43.1 \\
 & CodeV-Qwen1.5-7B & 77.6 & 88.2 & 90.7 & 52.7 & 62.5 & 67.3 & 36.6 & 53.3 & 61.3 \\
 & CodeV-Qwen2.5-7B & 77.3 & 87.9 & 90.1 & 57.9 & 66.7 & 69.7 & 39.3 & 63.5 & 74.2 \\
 & OriGen-7B & 74.1 & 82.4 & 85.7 & 54.4 & 60.1 & 64.2 & 50.6 & 68.3 & \textbf{74.3} \\
 & VeriPrefer-7B & 72.7 & 85.8 & - & 49.7 & 62.3 & - & 53.2 & 67.7 & - \\ \cmidrule(l){2-11} 
 & QiMeng-CRUX-SFT-7B (Ours) & 76.8 & 87.3 & \textbf{91.8} & 63.2 & \textbf{73.1} & \textbf{76.1} & 49.0 & 64.1 & 72.0 \\
 & QiMeng-CRUX-Final-7B (Ours) & \textbf{82.9} & \textbf{88.3} & 90.2 & \textbf{65.2} & 72.0 & 73.8 & \textbf{62.8} & \textbf{69.0} & 71.5 \\ \bottomrule
\end{tabular}
}
\caption{Main Results in VerilogEval-v1 and RTLLM-v1}
\label{tab: main_results_v1}
\end{table*}
\begin{table*}[ht]
\centering
\captionsetup{skip=7pt, belowskip=-8pt}
{\fontsize{9}{9.5}\selectfont  
\begin{tabular}{@{}ccccccccccc@{}}
\toprule
\multirow{2}{*}{Type} & \multirow{2}{*}{Model} & \multicolumn{3}{c}{VE-v2-CC (p@1)} & \multicolumn{3}{c}{VE-v2-SR (p@1)} & \multicolumn{3}{c}{RTLLM-V2} \\ \cmidrule(l){3-11} 
 &  & T=0 & T=0.8 & T-cross & T=0 & T=0.8 & T-cross & p@1 & p@5 & p@10 \\ \midrule
\multirow{2}{*}{\begin{tabular}[c]{@{}c@{}}Foundation\\ General Models\end{tabular}} & GPT-4o & 59.0 & 56.1 & 57.6 & 62.5 & 61.4 & 64.1 & 56.5 & 70.3 & 75.2 \\
 & Deepseek-V3-671B & 68.0 & 66.1 & 68.7 & 68.8 & 66.9 & 62.4 & 59.1 & 71.5 & 73.3 \\ \midrule
\begin{tabular}[c]{@{}c@{}}Foundation\\ Reasoning Models\end{tabular} & Deepseek-R1-671B & \textbf{82.7} & \textbf{81.0} & \textbf{79.1} & \textbf{83.3} & \textbf{79.8} & \textbf{77.5} & 64.7 & 75.8 & 79.7 \\ \midrule
\multirow{3}{*}{\begin{tabular}[c]{@{}c@{}}Verilog-Specific\\ Reasoning Models\end{tabular}} & HaVen-7B & - & - & - & - & - & 54.6 & - & - & - \\
 & CodeV-R1-Distill-7B & - & - & 65.6 & - & - & 65.2 & 57.2 & 71.9 & 77.1 \\
 & CodeV-R1-7B & - & - & 69.9 & - & - & 68.8 & \textbf{68.0} & \textbf{78.2} & \textbf{81.7} \\ \midrule
\multirow{3}{*}{\begin{tabular}[c]{@{}c@{}}General \\ Code Models\end{tabular}} & Qwen2.5-Coder-7B & 30.1 & 25.6 & 30.5 & 23.0 & 22.0 & 31.3 & 36.1 & 52.4 & 57.6 \\
 & Qwen2.5-Coder-32B & 44.2 & 41.5 & 46.6 & 46.8 & 41.6 & 47.5 & 47.8 & 63.9 & 67.8 \\
 & Deepseek-Coder-6.7B & 24.4 & 21.0 & - & 36.8 & 30.9 & - & 41.8 & 53.9 & 60.3 \\ \midrule
\multirow{5}{*}{\begin{tabular}[c]{@{}c@{}}Verilog-Specific \\ General Models\end{tabular}} & RTLCoder-6.7B & 35.9 & 31.5 & 33.7 & 36.8 & 30.9 & 31.1 & 33.6 & 45.3 & 49.2 \\
 & CodeV-Qwen2.5-7B & 58.3 & 49.9 & 60.4 & 44.8 & 37.4 & 42.2 & 41.0 & 60.1 & 68.1 \\
 & OriGen-7B & 49.3 & 47.2 & - & 49.3 & 46.8 & - & 50.9 & 60.9 & 64.0 \\ \cmidrule(l){2-11} 
 & QiMeng-CRUX-SFT-7B (Ours) & 64.7 & 58.9 & 65.5 & 59.6 & 57.0 & 61.1 & 52.3 & 68.6 & 73.4 \\
 & QiMeng-CRUX-Final-7B (Ours) & \textbf{68.0} & \textbf{66.7} & \textbf{67.6} & \textbf{64.7} & \textbf{64.4} & \textbf{64.2} & \textbf{63.8} & \textbf{70.6} & \textbf{73.9} \\ \bottomrule
\end{tabular}
}
\caption{Main Results in VerilogEval-v2 and RTLLM-v2}
\label{tab: main_results_v2}
\end{table*}

\section{Experiment}
\subsection{Experiment Settings}
\paragraph{Datasets}
Our dataset is derived from the CodeV \cite{codev} corpus, which was constructed by hierarchically parsing Verilog code with GPT to obtain (Description, Verilog Code) pairs. They used the Rouge-L metric to measure the similarity between the dataset and the benchmark, and removed samples with Rouge-L $>$ 0.5 to avoid data contamination. 
Following Corpus Categorization, we obtain about 40k Easy Questions, 18k Special Non-Text, and 107k Normal Data samples. During the Interface Degradation process, the complete interface is retained with a probability of 0.2, while all other variation elements are applied independently with a probability of 0.5. Only about 24k samples insert interface information in the middle of the description; the remaining samples include it only at the end.


\paragraph{Benchmarks}
To ensure a comprehensive evaluation, we adopt several representative and widely used Verilog generation benchmarks, including \textit{VerilogEval-V1} \cite{verilogeval-v1}, \textit{VerilogEval-V2} \cite{verilogeval-v2}, \textit{RTLLM-V1} \cite{rtllm-v1}, and \textit{RTLLM-V2} \cite{rtllm-v2}.
\textbf{VerilogEval-V1} consists of two sets of tasks: \textit{Machine} (143 tasks automatically generated by GPT) and \textit{Human} (156 hand-crafted tasks), both primarily aimed at evaluating code completion capabilities.
Building upon this, \textbf{VerilogEval-V2} further enhances the quality and diversity by introducing two subtasks: \textit{Code-Completion} that refined from V1-Human, and \textit{Spec-to-RTL} which imposes additional requirements on the model's ability to understand and generate core functional logic based on the specification. 
The \textbf{RTLLM} benchmark consists of four categories: \textit{Arithmetic}, \textit{Memory}, \textit{Control}, and \textit{Miscellaneous}, covering a wide range of challenges, including FSMs, combinational and sequential circuits, and long-text specifications. It poses a higher level of difficulty compared to VerilogEval. RTLLM-V2 expands upon RTLLM-V1, increasing the total number of tasks from 29 to 50, with V1 forming a subset of V2. Since some prior works only report results on RTLLM-V1, we also include our performance on V1 for fair comparison.

\paragraph{Training Setup}
We adopt Qwen2.5-Coder-7B-Instruct \cite{hui2024qwen2} as the backbone model and follow the proposed two-stage framework, applying supervised fine-tuning (SFT) for 2 epochs with \textit{LLaMaFactory} \cite{llamafactory} in Stage I, and CRUX-enhanced GRPO training for 1 epoch with \textit{EasyR1} \cite{easyr1} in Stage II.


\paragraph{Evaluation Setup}
Consistent with prior works, we adopt the \textit{pass@k} metrics to assess generation performance, estimating the probability that at least one correct solution is obtained among $k$ independent model outputs for each task, as defined in Equation \ref{equation:pass@k}. $n \geq k$ represents the number of independent solutions generated per task, and $c$ corresponds to how many of these trials are functionally correct.

\begin{equation}
\label{equation:pass@k}
\mathit{pass}@k := \mathbb{E}_{\text{problems}} \left[ 1 - \frac{\binom{n - c}{k}}{\binom{n}{k}} \right].
\end{equation}

Notably, the original \textit{VerilogEval-V2} paper requires reporting only \textit{pass@1} under two settings to highlight practical LLM evaluation: a low-temperature setting (T=0.0, top\_p=0.01, n=1) and a high-temperature setting (T=0.8, top\_p=0.95, n=20) \cite{verilogeval-v2}. 
However, many prior works still adopt a strategy of selecting the best performance among \textit{Temperature = 0.2, 0.5, 0.8}. To enable fair comparison, we also include \textit{pass@1} results under this strategy, denoted as \textit{T-cross}.
For the other benchmarks, we follow common practice and generate $n=20$ independent solutions per task to compute \textit{pass@1}, \textit{pass@5}, and \textit{pass@10}, reporting the best performance across T=0.2, 0.5, and 0.8.

\subsection{Main Results}
Our main experimental results are presented in Table \ref{tab: main_results_v1} and Table \ref{tab: main_results_v2}. We report the performance of Foundation Models \cite{deepseek-v3, qwen3technicalreport}, General Code Models \cite{hui2024qwen2, deepseek-coder}, and Verilog-Specific General Models \cite{rtlcoder, codev, origen, insightsfromverification}. Given the remarkable performance of Reasoning Models in recent years, we separately group Deepseek-R1 \cite{guo2025deepseek} and Verilog-Specific Reasoning Models \cite{haven, codev-r1}. 

It's worth mentioning that our approach is fundamentally different from Reasoning-based methods. Chain-of-Thought (CoT) reasoning encourages the model to reach the final answer through iterative “generate-reflect-revise” cycles across multiple thinking trajectories. In contrast, our method explicitly outputs a structured expression of the core intent and implementation details, with only one-shot generation to the final code. This distinction also leads to significant differences in token usage and computational cost: reasoning models typically require very long context windows of 16,384 tokens or more, whereas our model uses at most 4,096 tokens during both training and inference. Therefore, our approach is categorized as a Verilog-Specific General Model for performance comparison. Moreover, although CodeV-R1 shares the \textit{CodeV} name, it does not actually utilize the original CodeV dataset. Instead, it is built upon a newly constructed dataset from DeepSeek. Therefore, the improvements observed in CodeV-R1 and CRUX-V stem from different methods and data sources, and thus should not be directly compared.

For RTLLM-v1 and v2, we evaluate the performance of RTLCoder, CodeV-Qwen1.5, and CodeV-Qwen2.5. 
And for VerilogEval-V2, we evaluate General Code and Verilog-Specified Models under the required settings $1)$ T=0.00, top\_p=0.01, n=1 and $2)$ T=0.8, top\_p=0.95, n=20 and report their \textit{pass@1} results. The remaining results are derived from their own or prior works. Some entries in the table are left blank because the corresponding models are not open-sourced, or they were not reported in their papers.


The main results demonstrate our QiMeng-CRUX model achieves the best \textit{pass@1} performance across all benchmarks within both the General-Purpose Code Models and Verilog-Specific General Models, especially under challenging and realistic scenarios. It significantly outperforms GPT-4o on multiple tasks, most notably on RTLLM-v2, where it achieves a \textit{pass@1} of 63.8\%, surpassing the previous SOTA by 12.8\%, and outperforming Deepseek-V3, reaching performance comparable with Deepseek-R1.
On VerilogEval-v2, QiMeng-CRUX brings substantial improvements over previous SOTA models across nearly all metrics, such as from 58.3\% to 68.0\% (+9.7\%) in \textit{pass@1} on the CC task, and from 49.3\% to 64.7\% (+15.4\%) on the SR task. These results suggest that CRUX offers significant advantages in tackling complex code generation tasks and real-world design inputs.

Plus, QiMeng-CRUX demonstrates robust performance across diverse task settings and specification styles. Originating from the CodeV Datasets, QiMeng-CRUX consistently performs better across tasks with diverse input forms and exhibits stronger robustness to specification variations than CodeV2.5. It achieves balanced performance across CC and SR (68.0\% in CC and 64.7\% in SR), avoiding the large gap observed in CodeV2.5 (58.3\% in CC and 44.8\% in SR). Furthermore, QiMeng-CRUX achieves substantial improvements across all metrics on the RTLLM benchmark, highlighting the effectiveness and robustness of the CRUX methodology.

\begin{table*}[!t]
\centering
{\fontsize{9}{9.5}\selectfont  
\setlength{\tabcolsep}{4.5pt} 
\begin{tabular}{@{}lcccccccccccccc@{}}
\toprule
\multirow{2}{*}{Stage} & \multirow{2}{*}{Model} & \multicolumn{3}{c}{VE-v1-Machine} & \multicolumn{3}{c}{VE-v1-Human} & \multicolumn{2}{c}{VE-v2-CC (p@1)} & \multicolumn{2}{c}{VE-v2-SR (p@1)} & \multicolumn{3}{c}{RTLLM-V2} \\ \cmidrule(l){3-15} 
 &  & p@1 & p@5 & p@10 & p@1 & p@5 & p@10 & T=0 & T=0.8 & T=0 & T=0.8 & p@1 & p@5 & p@10 \\ \midrule
\multirow{3}{*}{Stage I} & CodeV & 77.3 & \textbf{87.9} & 90.1 & 57.9 & 67.6 & 69.7 & 58.3 & 49.9 & 44.8 & 37.4 & 41.0 & 60.1 & 68.1 \\
 & + RealSpec & \textbf{79.2} & 87.0 & 89.8 & 62.3 & 72.4 & 74.7 & 62.2 & 57.3 & 53.2 & 47.2 & 46.2 & 64.1 & 72.2 \\
 & + CRUX & 76.4 & 87.3 & \textbf{91.8} & \textbf{63.2} & \textbf{73.1} & \textbf{76.1} & \textbf{64.7} & \textbf{58.9} & \textbf{59.6} & \textbf{57.0} & \textbf{52.3} & \textbf{68.6} & \textbf{73.4} \\ \midrule
\multirow{2}{*}{Stage II} & \multicolumn{1}{l}{w/o CRUX-Reward} & 80.6 & 87.2 & \textbf{91.4} & 61.7 & 69.3 & 73.1 & 61.5 & 59.5 & 62.8 & 58.3 & 59.5 & \textbf{73.0} & \textbf{74.8} \\
 & \multicolumn{1}{l}{w CRUX-Reward} & \textbf{82.9} & \textbf{88.3} & 90.2 & \textbf{65.2} & \textbf{72.0} & \textbf{73.8} & \textbf{68.0} & \textbf{66.7} & \textbf{64.7} & \textbf{64.4} & \textbf{63.8} & 70.6 & 73.9 \\ \bottomrule
\end{tabular}
}
\caption{Ablations of Both Stages}
\label{tab: ablation_two_stages}
\vspace{-3pt}
\end{table*}

\begin{figure}[t]
    \captionsetup{skip=3pt, belowskip=-10pt}
    \centering
    \includegraphics[width=0.48\textwidth]{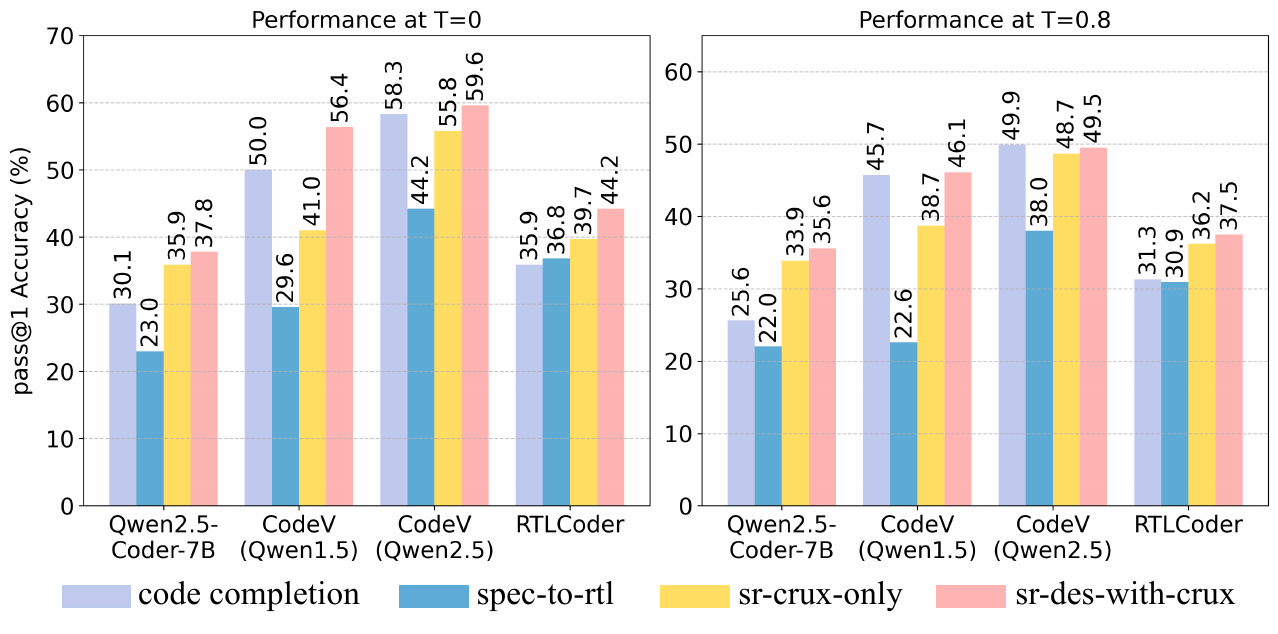}
    \caption{Using CRUX alone (crux\_only) already leads to notable gains compared to using the original specification directly. Further combining CRUX with the original descriptions (des\_with\_crux) yields the best performance, even outperforming the Code-Completion task in most cases.}
    \label{fig:crux-works}
    \vspace{-3pt}
\end{figure}




\subsection{Ablation Study}
In this section, we present a systematic ablation study of our CRUX framework within each of the two stages to assess the effectiveness of our methods. Plus, we investigate the transferability of the learned CRUX space to some general-purpose and verilog-specific code models to further evaluate its quality. The results are presented in Table \ref{tab: ablation_two_stages} and Figure \ref{fig:crux-works}.


\paragraph{Ablation of RealSpec and CRUX in Stage I}
We focus on evaluating the impact of RealSpec and CRUX of Stage I. The experimental groups are configured as:
(i) CodeV: SFT on (Description, Verilog) dataset (original CodeV form);
(ii) CodeV + RealSpec: SFT on (RealSpec, Verilog) dataset;
(iii) CodeV + RealSpec + CRUX: SFT on (RealSpec, CRUX, Verilog) dataset.
Results in Stage I of Table \ref{tab: ablation_two_stages} demonstrate the benefits of RealSpec and CRUX. Compared to the baseline CodeV, incorporating RealSpec alone leads to consistent improvements across almost all benchmarks, especially on the Spec-to-RTL, with \textit{pass@1} improving significantly by 8.4\% in T=0 and by 9.8\% in T=0.8, indicating simulation of realistic input distributions enhances the model’s robustness and generalization to real-world usage patterns.  
Further introducing CRUX yields the best overall performance in Stage I, bringing at least a 5\% improvement in \textit{pass@1} on both the SR and RTLLM-v2 benchmarks. Notably, RealSpec alone brings limited gains on the RTLLM, while CRUX continues to deliver clear improvements even under these more challenging settings, demonstrating its ability to capture core design intent and guide the model toward more precise Verilog generation. In summary, RealSpec improves robustness under natural prompt variation, and CRUX excels in tasks that require a deep understanding of core design intent and careful attention to implementation details. 

\paragraph{Ablation of CRUX-Enhancement in Stage II}

We focus on evaluating the impact of CRUX-Reward used in GRPO of Stage II by comparing models trained with and without CRUX-Reward. As shown in Stage II of Table \ref{tab: ablation_two_stages}, introducing CRUX-Reward leads to a notable improvement in \textit{pass@1}, while a slight drop in \textit{pass@10}. A similar trend can also be observed if compared with SFT models. While RL-based methods are known to enhance generation precision, they often constrain the diversity of candidate solutions. In particular, compared to GRPO without CRUX-Reward, the incorporation of CRUX-Reward further encourages the model to optimize toward more compact solution spaces that ensure both higher-quality CRUX and more precise Verilog code. This observation highlights the role of CRUX not only as a structured expression but also as an inductive bias that shapes the model’s generation space toward more precise yet less diverse outputs.

\paragraph{Transferability of the Learned CRUX Space}
Beyond evaluating our model's performance on Verilog benchmarks, we also investigate whether the learned CRUX space provides transferable benefits to general-purpose code models to assess its quality.
Following the setup in Figure 1, we also conduct experiments on the Spec-to-RTL task to evaluate the guiding effect of CRUX as a prompt for general-purpose code models.
Specifically, we test two settings: (i) using CRUX alone as the prompt, and (ii) appending CRUX as a bridge to the original descriptions.
As shown in Figure 5, we observe consistent and sustained performance improvements under both settings, which reflect the quality of the learned CRUX space itself and further highlight its effectiveness in bridging natural language descriptions and precise code generation.
These results confirm that the CRUX not only benefits our own model but also provides meaningful guidance to general-purpose or Verilog-specified code models, even without any additional training.

\section{Conclusion}
In this work, we present CRUX, a structured intermediate space designed to bridge the gap between free-form natural language descriptions and formal Verilog code. By explicitly modeling \textit{Module Interfaces}, \textit{Core Functions}, and \textit{Key Considerations}, CRUX captures essential design intent and provides clear, targeted guidance for downstream code generation. Through a two-stage framework combining \textit{Joint Expression Modeling} and \textit{Dual-Space Optimization}, our method not only improves generation accuracy but also enhances the robustness and interpretability of large language models in hardware design tasks. Extensive evaluations show that QiMeng-CRUX achieves state-of-the-art performance across multiple Verilog benchmarks, demonstrating the effectiveness of semantically structured guidance in enabling precise and synthesizable HDL generation.


\section{Acknowledgments}
This work is partially supported by the NSF of China (Grants No.62341411, 62525203, U22A2028, 62222214), Strategic Priority Research Program of the Chinese Academy of Sciences (Grants No.XDB0660200, XDB0660201, XDB0660202), CAS Project for Young Scientists in Basic Research (YSBR-029) and Youth Innovation Promotion Association CAS.

\bibliography{aaai2026}
\appendix
\section{Appendix}
\subsection{Details of CRUX-Enhanced GRPO}
\paragraph{Preliminary of GRPO}
Group Relative Policy Optimization (GRPO) is an \textit{on-policy} reinforcement learning algorithm built upon the Proximal Policy Optimization (PPO) framework. GRPO removes the value model to significantly reduce inference cost, while introducing \textit{group relative advantage estimation} to more accurately assess the quality of model outputs. Furthermore, a KL-divergence penalty is incorporated to stabilize policy updates and prevent the policy from deviating excessively against the reference model.

In details, for a given query \(q\), GRPO samples a group of \(G\) outputs
\[
\{o_i\}_{i=1}^{G} \sim \pi_{\theta_{\mathrm{old}}}(\cdot \mid q),
\]
and evaluates each output \(o_i\) with a reward function to obtain rewards \(\{r_i\}_{i=1}^{G}\). The group-relative advantage for each sample is obtained by standardizing the group rewards:
\[
A_i \;=\; \frac{r_i - \operatorname{mean}(\mathbf{r})}{\operatorname{std}(\mathbf{r})},
\qquad \text{where }\; \mathbf{r}=\{r_i\}_{i=1}^{G}.
\]

To stabilize training and prevent the updated policy \(\pi_\theta\) from deviating excessively from a reference policy \(\pi_{\mathrm{ref}}\), GRPO incorporates a KL-divergence penalty. Define the importance sampling ratio for each token \(o_{i,t}\) in response sequence \(o_i\) (given query \(q\) and preceding tokens \(o_{i,<t}\)) as:
\[
r_{i,t}(\theta) = \frac{\pi_\theta(o_{i,t} \mid q, o_{i,<t})}{\pi_{\theta_{\mathrm{old}}}(o_{i,t} \mid q, o_{i,<t})}.
\]
The GRPO objective is formulated with clipped importance weighting and a KL penalty, averaged over groups and response tokens, as shown in Eq.\ref{eq:grpo_final}.

Here, \(\epsilon>0\) denotes the PPO clipping threshold and \(\beta\ge 0\) regulates the KL penalty strength, while \(\hat{A}_{i,t}\) represents the token-level advantage estimate for the \(t\)-th token in the \(i\)-th output sequence. In practice, the expectation in Eq. \eqref{eq:grpo_final} is approximated by sampling \((q,a)\) pairs from the data distribution \(\mathcal{D}\), and for each query \(q\), drawing \(G\) output sequences \(\{o_i\}\) from the old policy \(\pi_{\theta_{\mathrm{old}}}(\cdot\mid q)\). The double averaging over \(G\) sequences (via \(1/G\)) and token positions (via \(1/|o_i|\)) ensures stable gradient estimation across variable-length outputs and group-level comparisons.

\begin{figure*}[!t]
\centering
\begin{align}
\mathcal{J}_{\mathrm{GRPO}}(\theta)
&= \mathbb{E}_{(q,a)\sim\mathcal{D},\, \{o_i\}_{i=1}^{G}\sim\pi_{\theta_{\mathrm{old}}}(\cdot\mid q)}
\nonumber
\\
\Bigg[ 
&\quad \frac{1}{G}\sum_{i=1}^{G} \frac{1}{|o_i|}
\sum_{t=1}^{|o_i|}
\Big(
\min\big(r_{i,t}(\theta)\,\hat{A}_{i,t},\;
\operatorname{clip}(r_{i,t}(\theta),\,1-\epsilon,\,1+\epsilon)\,\hat{A}_{i,t}\big)
-\, \beta\, D_{\mathrm{KL}}\!\left(\pi_{\theta}\,\|\,\pi_{\mathrm{ref}}\right)
\Big)
\Bigg].
\label{eq:grpo_final}
\end{align}
\end{figure*}

\begin{figure}[t]
  \centering
  \includegraphics[width=0.47\textwidth]{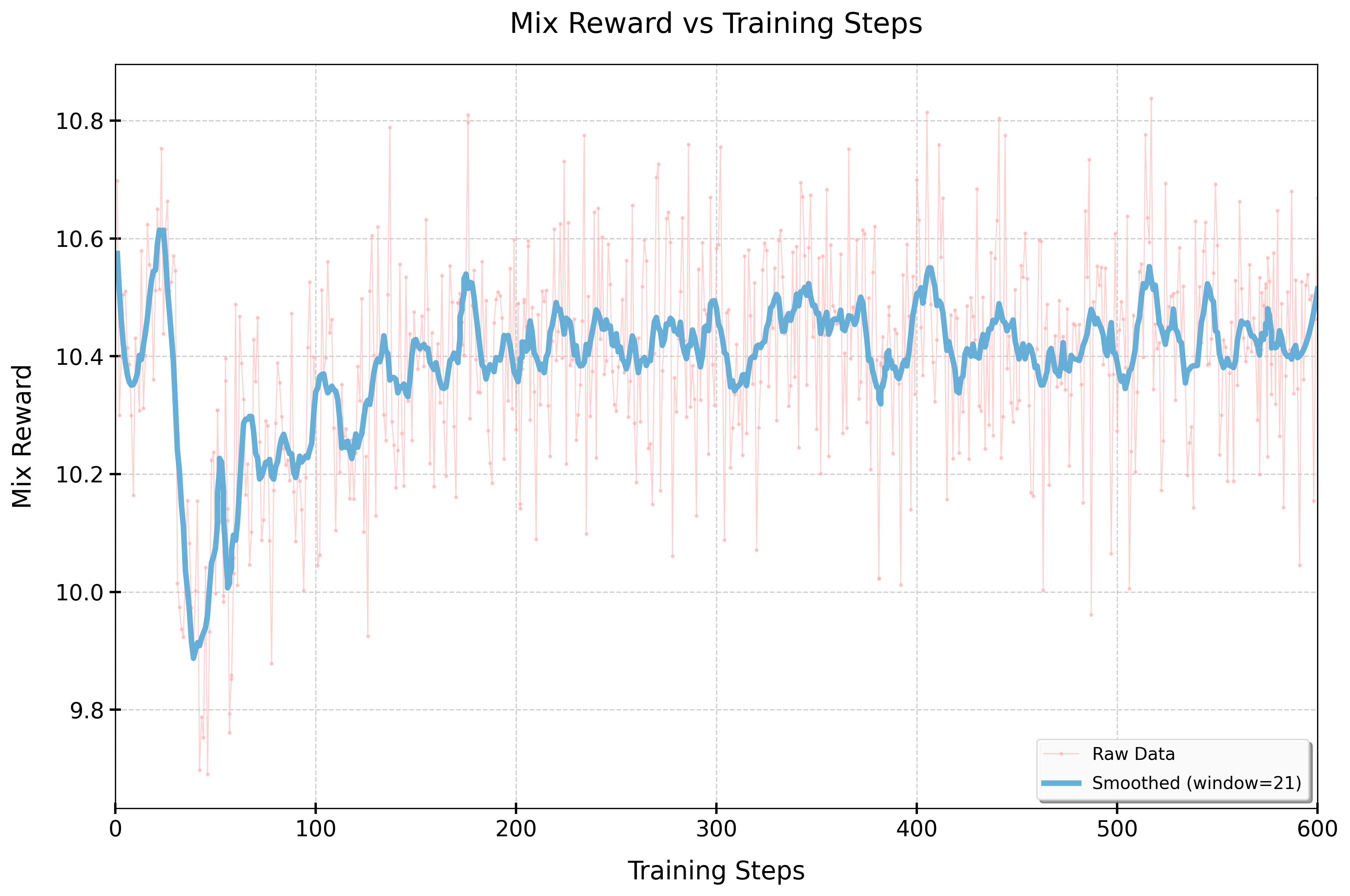} 
  \caption{Mixed reward curve over training steps in GRPO. The curve shows how the overall reward changes with training steps under different reward weight schedules.}
  \label{fig:mix_reward_curve}
\end{figure}

\begin{figure}[h]
  \centering
  \includegraphics[width=0.47\textwidth]{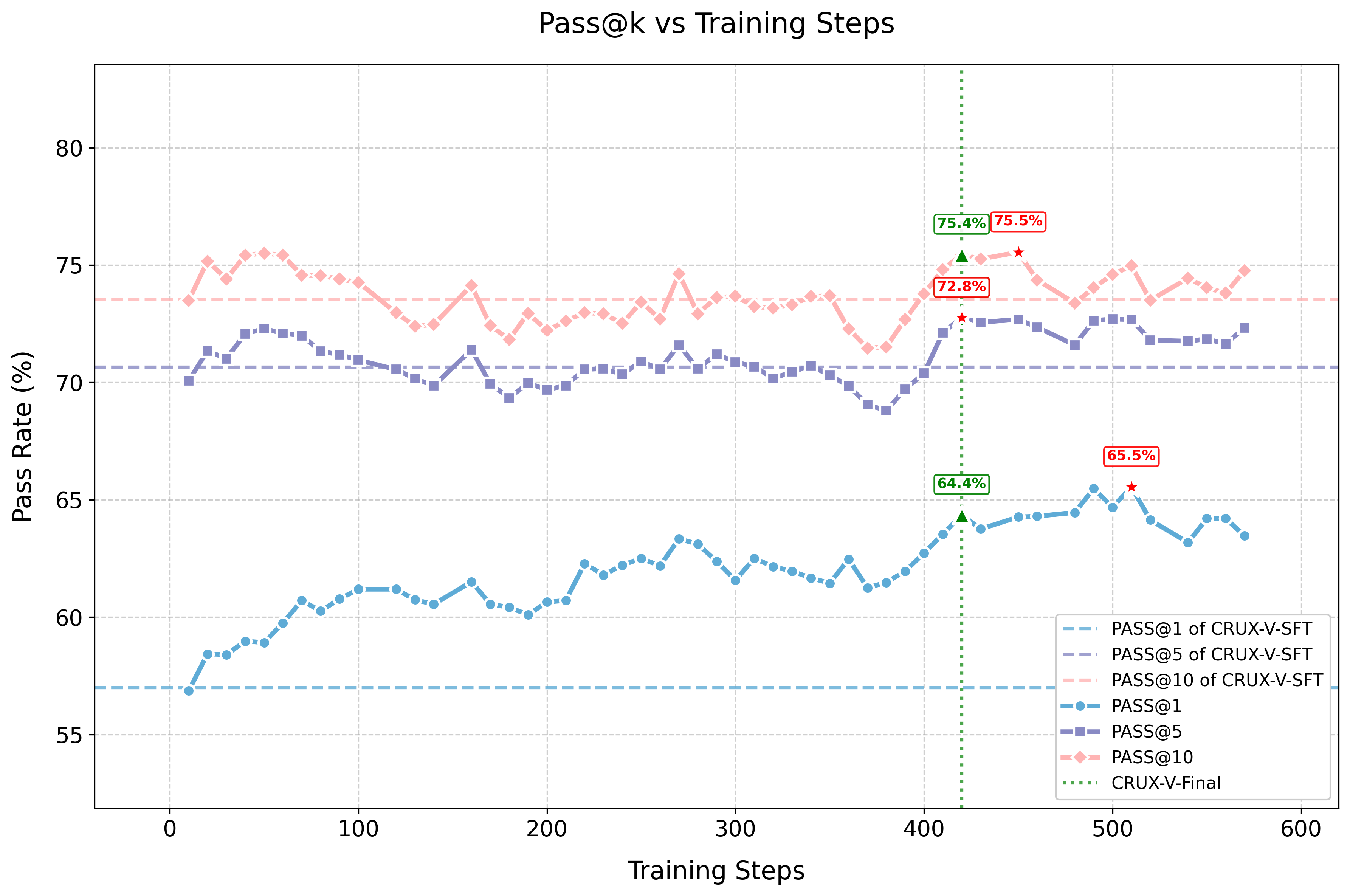} 
  \caption{Evolution of the \textit{pass@k} metrics on the Spec-to-RTL benchmark during GRPO training. Models are evaluated every 10 steps on the benchmark.}
  \label{fig:sr_passk_by_step}
\end{figure}

\paragraph{CRUX-Enhanced GRPO}
We have made certain improvements to GRPO to adapt it to the CRUX method. Compared to the original GRPO optimization objective, we have implemented two key modifications: 1) Removing the KL Divergence; 2) Progressive Multi-Reward Mixing.

\textbf{1) Removing KL Divergence.} QiMeng-CRUX constructs an intermediate space that bridges free-form natural language descriptions and domain-specific Verilog code. To ensure the effectiveness of this space for precise Verilog generation, we introduce a CRUX-Reward, which directly optimizes the semantic alignment between the CRUX and correct Verilog codes. Consequently, we do not employ KL-divergence to regularize the policy model. Removing the KL constraint encourages the model to generate more diverse CRUXes and Verilog codes, while CRUX-Reward and Code-Reward respectively serve as the primary objectives that guide the generation toward correctness and faithfulness, providing a more direct and effective optimization mechanism.

To further validate the effectiveness of this design, we conduct an ablation study based on QiMeng-CRUX-SFT, examining the impact of removing the KL-divergence term. The evaluation is performed on three benchmarks: VerilogEval-V1-Human, VerilogEval-V2-SR, and RTLLM-v2, and we report \textit{pass@1} under identical training steps and datasets. The results are summarized as follows.

\begin{table}[h]
\begin{tabular}{@{}cccc@{}}
\toprule
Models & Human & Spec-to-RTL & RTLLM-v2 \\ \midrule
CRUX-V-SFT(base) & 63.2 & 59.6 & 52.3 \\
w KL-Penalty & 63.8 & 62.1 & 59.7 \\
w/o KL-Penalty & 65.2 & 64.7 & 63.8 \\ \bottomrule
\end{tabular}
\caption{Ablation study on the KL-divergence penalty in Stage II of QiMeng-CRUX training across three benchmarks, showing consistent performance gains when the KL penalty is removed.}
\label{tab:ablation_of_kl}
\end{table}

As shown in Table \ref{tab:ablation_of_kl}, removing the KL-divergence penalty leads to a noticeable improvement across the evaluated benchmarks. These results indicate that the KL constraint limits the expressiveness of the policy model, suppressing the generation of diverse yet semantically aligned CRUX space. In contrast, optimizing solely with CRUX-Reward and Code-Reward without KL-divergence provides a more flexible and task-aligned training signal, leading to more faithful Verilog generation.

\textbf{2) Progressive Multi-Reward Mixing. }
To further stabilize training and ensure early-stage code usability, we incorporate two additional auxiliary rewards: (i) \textit{Format-Reward}, which encourages generation with essential structural elements required by CRUX; and (ii) \textit{Compile-Reward}, which ensures that generated code is at least syntactically compilable before functional testing. To balance these rewards over training, we adopt a progressive weighting strategy that sets the weights of Format-, Compile-, CRUX- and Code-Reward at [1, 3, 4, 6] initially and updates to [0.5, 1.5, 4, 8] after the first 0.1 epoch (so the full score is 14). This schedule encourages the model to produce well-formatted CRUX and compilable code in the early training phase, while shifting the focus toward improving function correctness in later steps, creating a smoother optimization landscape for reinforcement learning.

\begin{figure*}[!t]
  \centering
  \includegraphics[width=\textwidth]{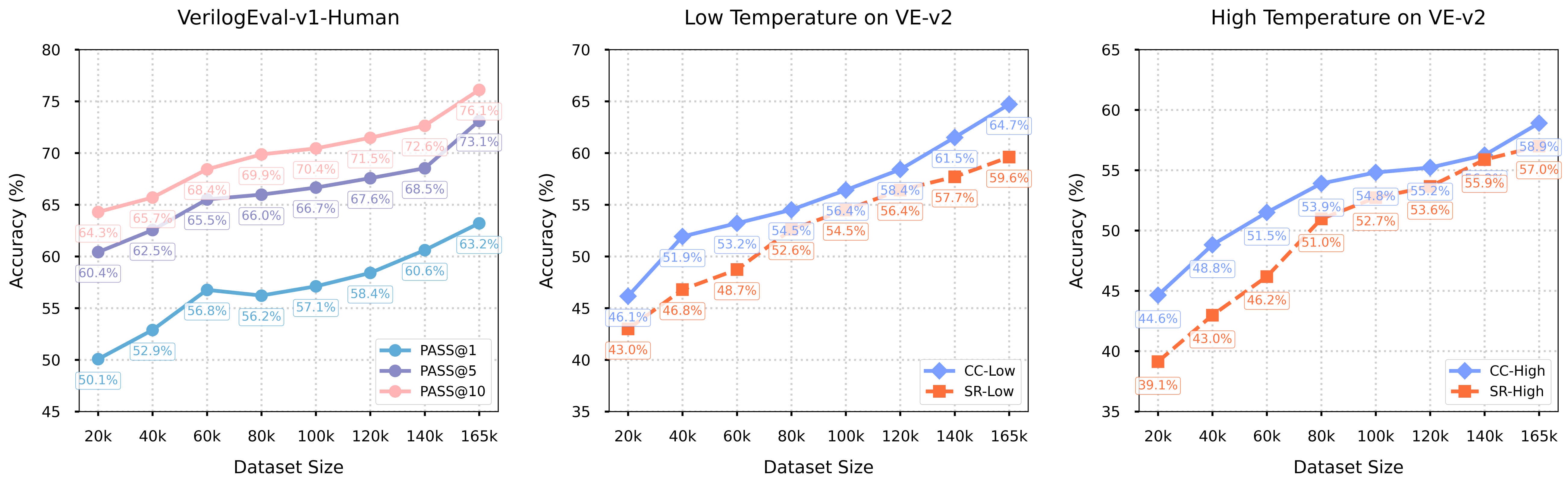} 
  \caption{Ablation of Dataset Scaling. We randomly sampled datasets of different scales to train SFT, and obtained their performance on the VerilogEval-V1-Human task, as well as the VerilogEval-V2 Code-Completion and Spec-to-RTL tasks.}
  \label{fig:data_scaling}
\end{figure*}

\paragraph{Dynamics of GRPO training}

Figure\ref{fig:mix_reward_curve} illustrates the evolution of the overall reward during the GRPO training process. At the early stages of training, the reward exhibits a sharp decline, which can be attributed to fluctuations in the difficulty of randomly partitioned data batches and the model’s insufficient adaptation after SFT-based static data training. However, the reward quickly increases thereafter, as the reward weights are initially set to ([1, 3, 4, 6]), emphasizing relatively easier objectives, learning the correct output format and generating compilable code, which enables the model to achieve rapid performance gains in the early training stage.
After around 20–30 training steps, the reward weights are adjusted to 
[0.5,1.5,4,8], leading to a noticeable drop in reward. Subsequently, the training proceeds normally, with the reward gradually improving and stabilizing after approximately 400 steps.

During GRPO training, we save a model checkpoint every 10 steps. Figure\ref{fig:sr_passk_by_step} illustrates the evolution of the \textit{pass@k} metric on the Spec-to-RTL benchmark from VerilogEval-V2 as training progresses. As shown, \textit{pass@1} steadily increases over time but experiences a slight drop after around 460 steps, while \textit{pass@10} decreases initially and then shows a mild improvement after 400 steps. The red star marks the best performance among the steps, and the green dashed line indicates the final model we selected. We adopt the checkpoint at 420 steps as our QiMeng-CRUX-Final model—not the best on Spec-to-RTL alone, but demonstrating the most balanced performance across multiple benchmarks.

\subsection{Additional Experiments and Supplementary}
\paragraph{Parameters and Computational Usage}
Some hyperparameters used in training and computational usage are listed in Table\ref{tab:training_paras}.
\begin{table}[!h]
\begin{tabular}{@{}lll@{}}
\toprule
Training Method & SFT & GRPO \\ \midrule
epoch & 2 & 1 \\
learning\_rate & 1.0e-5 & 1.0e-6 \\
strategy & adam & adamw \\
warmup\_ratio & 0.1 & 0.0 \\
rollout\_n & - & 5 \\
rollout\_batch & - & 512 \\
max\_length & 4096 & 4096 \\
n\_gpu & 4$\times$A100,80g & 4$\times$A100,80g \\
time cost & 30h & 36h \\
temperature & - & 1.0 \\
top-p & - & 0.99 \\ \bottomrule
\end{tabular}
\caption{Training parameters and computational usage.}
\label{tab:training_paras}
\end{table}

\begin{table*}[]
\centering
\begin{tabular}{@{}cccccccccc@{}}
\toprule
\multirow{2}{*}{Models} & \multirow{2}{*}{Prompt} & \multicolumn{3}{c}{VE-v1-Human} & \multicolumn{2}{c}{VE-v2-SR(p@1)} & \multicolumn{3}{c}{RTLLM-V2} \\ \cmidrule(l){3-10} 
 &  & p@1 & p@5 & p@10 & T=0 & T=0.8 & p@1 & p@5 & p@10 \\ \midrule
\multirow{3}{*}{Qwen2.5-Coder-7B} & Original & 22.9 & 36.0 & 39.5 & 23.0 & 22.0 & \textbf{36.1} & 52.4 & 57.6 \\
 & Only CRUX & 33.1 & 44.2 & 49.7 & 35.9 & 33.9 & 32.4 & 50.0 & 59.7 \\
 & Des + CRUX & \textbf{34.6} & \textbf{45.9} & \textbf{50.2} & \textbf{37.8} & \textbf{35.6} & 33.9 & \textbf{54.6} & \textbf{61.5} \\ \midrule
\multirow{4}{*}{RTLCoder} & Original(in paper) & \textit{\textbf{41.6}} & \textit{\textbf{50.1}} & \textit{\textbf{53.4}} & - & - & 33.6 & 45.3 & 49.2 \\
 & \multicolumn{1}{l}{Original(vllm infer)} & \multicolumn{1}{l}{\textit{36.5}} & \multicolumn{1}{l}{\textit{45.7}} & \textit{48.8} & 36.8 & 30.9 & - & - & - \\
 & Only CRUX & 37.1 & 45.0 & 49.4 & 39.7 & 36.2 & 32.9 & 41.7 & 45.9 \\
 & Des + CRUX & \textbf{38.7} & \textbf{46.8} & \textbf{51.3} & \textbf{44.2} & \textbf{37.5} & \textbf{36.2} & \textbf{48.4} & \textbf{56.2} \\ \midrule
\multirow{3}{*}{CodeV-Qwen1.5} & Original & 52.7 & 62.5 & 67.2 & 29.6 & 22.6 & 26.0 & 49.5 & 61.2 \\
 & Only CRUX & 54.3 & 65.5 & 69.7 & 41.0 & 38.7 & 32.0 & 50.0 & 55.7 \\
 & Des + CRUX & \textbf{59.6} & \textbf{68.2} & \textbf{71.8} & \textbf{56.4} & \textbf{46.1} & \textbf{38.5} & \textbf{57.7} & \textbf{66.7} \\ \midrule
\multirow{3}{*}{CodeV-Qwen2.5} & Original & 57.9 & 66.7 & 69.7 & 44.2 & 38.0 & 41.0 & 60.1 & 68.1 \\
 & Only CRUX & 59.6 & 69.5 & 72.6 & 55.8 & 48.7 & 43.0 & 61.9 & 68.7 \\
 & Des + CRUX & \textbf{62.2} & \textbf{72.5} & \textbf{74.7} & \textbf{59.6} & \textbf{49.5} & \textbf{49.9} & \textbf{64.5} & \textbf{71.6} \\ \bottomrule
\end{tabular}
\caption{Transferability of the learned CRUX space across benchmarks. Integrating CRUX consistently improves performance across models. For RTLCoder, both the original paper results and our reproduced results (using vLLM inference) are reported, as slight deviations may stem from differences in inference frameworks, precision handling, or decoding configurations.}
\label{tab:transferability}
\end{table*}

\paragraph{Ablation of Dataset Scaling}

In addition to the results presented in the main results section, we have also conducted experiments on dataset scaling to assess how model performance improves with increasing dataset size. Specifically, we sampled datasets of different sizes from the original dataset, and the results of the tests on the benchmarks are shown in Figure \ref{fig:data_scaling}.
 As shown in the plot, all three tasks, VerilogEval-v1-Human, VerilogEval-v2 Code Completion (CC) and Spec-to-RTL (SR), exhibit a consistent upward trend as the dataset grows from 20k to 165k. Accuracy improves steadily across different difficulty levels and evaluation metrics, demonstrating that larger datasets consistently lead to better model performance.

\paragraph{Transferability of the Learned CRUX Space}
In the main results, we demonstrate that the CRUX space produced by QiMeng-CRUX-Final on the Spec-to-RTL benchmark can serve either as a complete prompt or as a bridging component to the original design prompt. This shows that the learned CRUX space provides transferable benefits to general-purpose code models, without requiring any additional training. To further support this observation, we evaluate the transferability of CRUX on two additional benchmarks that also reflect real-world hardware design scenarios: VerilogEval-V1-Human, which features hand-crafted and semantically diverse prompts, and RTLLM-V2, which involves high-level and complex specifications resembling practical workflows. Full results are presented in Table~\ref{tab:transferability}.

Across all models and evaluation settings, we observe consistent improvements when incorporating CRUX, either as a standalone prompt or when combined with the original design prompt (Des + CRUX). In particular, the Des + CRUX configuration yields the strongest performance overall, suggesting that the structured CRUX space effectively bridges the gap between natural language descriptions and precise Verilog generation. Rather than serving merely as additional context, CRUX acts as a semantically aligned intermediate space that helps general-purpose models better interpret and operationalize hardware-specific design intent and implementation constraints.

While reproducing the baseline results, we observed that the performance of RTLCoder on the VerilogEval-V1-Human benchmark deviates slightly from the numbers reported in the original paper. This discrepancy is likely attributed to differences in inference frameworks (e.g., vLLM we used), precision handling, or decoding configurations such as sampling strategies and random seeds. For transparency, we include both the original and reproduced results in Table~\ref{tab:transferability}. Importantly, despite these variations, the performance gains introduced by CRUX remain consistent and substantial, highlighting its robustness across realistic hardware design tasks.

\subsection{Examples of CRUX on benchmarks}
To more intuitively illustrate the learned CRUX, we sampled task descriptions and their corresponding CRUX from the benchmarks, and presented the pass rates (with n=20) of three models.

\paragraph{VerilogEval-V1 (e.g. \lstinline{ece241\_2013\_q8})}

\begin{table}[b]
\begin{tabular}{@{}ccc@{}}
\toprule
ece241\_2013\_q8 & Machine-passrate & Human-passrate \\ \midrule
Qwen2.5-Coder-7B & 0.30 & 0.00 \\
CodeV-Qwen2.5 & 0.75 & 0.30 \\
QiMeng-CRUX & 1.00 & 0.60 \\ \bottomrule
\end{tabular}
\end{table}

\paragraph{Description in Machine}
	When the input x is 0, and the state is S, the state next is S. 
	when the input x is 1, and the state is S, the state next is S1.
	when the input x is 0, and the state is S1, the state next is S10. 
	when the input x is 1, and the state is S1, the state next is S1.
	when the input x is 0, and the state is S10, the state next is S. 
	when the input x is 1, and the state is S10, the state next is S1.
	when the input state is S, the output z is 0.
	when the input state is S1, the output z is 0.
	when the input state is S10, the output z is the input x.

\paragraph{Description in Human} Implement a Mealy-type finite state machine that recognizes the sequence "101" on an input signal named x. Your FSM should have an output signal, z, that is asserted to logic-1 when the "101" sequence is detected. Your FSM should also have a negative edge triggered asynchronous reset. You may only have 3 states in your state machine. Your FSM should recognize overlapping sequences.

\paragraph{CRUX} 
\textit{Module Interface}
\begin{lstlisting}[style=Verilog]
module top_module (
    input clk,
    input aresetn,
    input x,
    output reg z
);
\end{lstlisting}
\textit{Core Functions}:
Design a finite state machine that transitions between three states S, S1, and S10 based on the input x. The output z should be 0 when the state is S or S1, and it should be equal to the input x when the state is S10. The FSM should have an asynchronous reset signal aresetn to reset the state to S.

\begin{itemize}
    \item From S:
    \begin{itemize}
        \item On input x = 0, stay in S.
        \item On input x = 1, transition to S1.
    \end{itemize}
    
    \item From S1:
    \begin{itemize}
        \item On input x = 0, transition to S10.
        \item On input x = 1, stay in S1.
    \end{itemize}
    
    \item From S10:
    \begin{itemize}
        \item On input x = 0, transition to S.
        \item On input x = 1, transition to S1.
    \end{itemize}
\end{itemize}

\textit{Key Considerations}:
The state machine has three states: S, S1, and S10. The transitions between these states are determined by the input x. The output z is 0 when the state is either S or S1, and it is equal to the input x when the state is S10. The state machine also includes an asynchronous reset signal aresetn, which resets the state to S when active (low).

\paragraph{VerilogEval-V2 (e.g. \lstinline{dff8p})} 

\begin{center}
\centering
\begin{tabular}{@{}ccc@{}}
\toprule
dff8p & CC-passrate & SR-passrate \\ \midrule
Qwen2.5-Coder-7B & 0.05 & 0.15 \\
CodeV-Qwen2.5 & 0.85 & 0.10 \\
QiMeng-CRUX & 1.00 & 1.00 \\ \bottomrule
\end{tabular}
\end{center}

\paragraph{Description in Code-Completion}
Create 8 D flip-flops with active high synchronous reset. The flip-flops must be reset to 0x34 rather than zero. All DFFs should be triggered by the negative edge of clk.

\begin{lstlisting}[style=Verilog]
module TopModule (
  input clk,
  input [7:0] d,
  input reset,
  output reg [7:0] q
);
\end{lstlisting}

\paragraph{Description in Spec-to-RTL}
I would like you to implement a module named TopModule with the following
interface. All input and output ports are one bit unless otherwise
specified.

\begin{itemize}[label=-]
    \item input clk
    \item input reset
    \item input d (8 bits)
    \item output q (8 bits)
\end{itemize}

Implement a module that includes 8 D flip-flops with active high synchronous reset. The flip-flops must be reset to 0x34 rather than zero. All DFFs should be triggered by the negative edge of clk.

\paragraph{CRUX} 
\textit{Module Interface}

\begin{lstlisting}[style=Verilog]
module TopModule (
    input clk,
    input reset,
    input [7:0] d,
    output reg [7:0] q
);
\end{lstlisting}

\textit{Core Function}
\begin{itemize}[label=-]
    \item Implements 8 D flip-flops with synchronous reset  
    \item Reset value is 0x34
    \item Negative edge-triggered
\end{itemize}


\newpage

\paragraph{RTLLM (e.g. \lstinline{clkgenerator.v})}

\begin{center}
\centering
\begin{tabular}{@{}cc@{}}
\toprule
RTLLM-clkgenerator & passrate \\ \midrule
Qwen2.5-Coder-7B & 0.00 \\
CodeV-Qwen2.5 & 0.05 \\
QiMeng-CRUX & 1.00 \\ \bottomrule
\end{tabular}
\end{center}

\paragraph{Description}
Please act as a professional Verilog designer.
A clock generator module that produces a periodic clock signal, toggling its output state at regular intervals defined by the PERIOD parameter.

\noindent
Module name:\\
\hspace*{1em} clkgenerator

\vspace{0.5em}
\noindent
Parameter:\\
\hspace*{1em} PERIOD = 10

\vspace{0.5em}
\noindent
Output ports: \\
\hspace*{1em} clk: The output clock signal generated by the module.
\vspace{0.5em}
\noindent
Implementation: \\
\hspace*{1em} This module uses an initial block to set the initial state of the clock signal to 0. The clock toggles every half of the specified PERIOD, creating a square wave clock signal with the desired frequency.
Give me the complete code.

\paragraph{CRUX} 
\textit{Module Interface}
\begin{lstlisting}[style=Verilog]
module clkgenerator #(
    parameter PERIOD = 10
)(output reg clk);
\end{lstlisting}

\textit{Core Function}
\begin{itemize}[label=-]
    \item Implements clock generation with configurable period 
    \item Produces square wave output  
    \item Initializes clock state to 0 
    \item Toggles output every half period
\end{itemize}

From the above examples, it can be observed that CRUX is more concise and accurate. It presents useful information in a more intuitively structured manner, thereby supporting precise code generation for models.
\end{document}